\documentclass[10pt,twocolumn,letterpaper]{article}

\usepackage[pagenumbers]{wacv} 

\newcommand{\TODO}[1]{\textbf{\color{red}[TODO: #1]}}

\renewcommand{\TODO}[1]{}

\usepackage{tcolorbox}
\tcbuselibrary{breakable}
\usepackage{placeins}
\usepackage{multirow}
\usepackage{booktabs}
\usepackage{graphicx}
\usepackage{booktabs}
\usepackage{tabularx}

\usepackage{marginnote}
\newcommand{\xiaofan}[1]{%
  \marginnote{\scriptsize\color{red}#1}%
}
\renewcommand{\xiaofan}[1]{}

\definecolor{wacvblue}{rgb}{0.21,0.49,0.74}
\usepackage[pagebackref,breaklinks,colorlinks,allcolors=wacvblue]{hyperref}

\def\wacvPaperID{2456} 
\def\confName{WACV}
\def\confYear{2027}

\title{From Vision to Harvest: Benchmarking Vision-Language Models for Multi-Arm Robotic Fruit Harvesting}

\author{
Vrishan Inukollu\textsuperscript{*1}, 
Adyan Zaman\textsuperscript{*1},
Anvi Kudaraya\textsuperscript{*1},
Carlos Lazcano\textsuperscript{1},
Yuankai Zhu\textsuperscript{2}, \\
Stavros Vougioukas\textsuperscript{2},
Xiaofan Yu\textsuperscript{1}\\
($^{*}$Equal contribution) \\
\textsuperscript{\rm 1} University of California, Merced,
\textsuperscript{\rm 2} University of California, Davis \\
{\tt\small \{vinukollu, azaman7, akudaraya, clazcano, xiaofanyu\}@ucmerced.edu} \\
{\tt\small \{ykzhu, svougioukas\}@ucdavis.edu} \\
}

\begin{document}
\maketitle
\begin{abstract}
Multi-arm robotic harvesting offers a promising path to improve harvesting efficiency and reduce reliance on manual labor. However, practical deployment remains challenging because the system must generalize across diverse environments while efficiently coordinating multiple arms in a shared workspace. Existing methods often require substantial data collection in target environments or rely on simplifying assumptions that limit planning quality. In this work, we introduce the first comprehensive benchmark for evaluating pretrained Vision-Language Models (VLMs) on zero-shot multi-arm fruit harvesting planning. Our benchmark uses real-world apple and citrus orchard images and compares a VLM-based planning pipeline with a traditional perception-and-planning pipeline. The VLM pipeline directly generates harvesting sequences and waypoints for each arm, while a lightweight trajectory verifier checks for collisions. Our results show that frontier VLMs can generate effective multi-arm harvesting plans zero-shot, but a practical deployment remains limited by accurate 3D waypoint generation and collision-aware coordination. These results highlight both the promise and current limitations of pretrained VLMs for multi-arm robotic harvesting.
\end{abstract}

\section{Introduction}
\label{sec:intro}

\begin{figure}[t]
\centering
\includegraphics[width=1\linewidth]{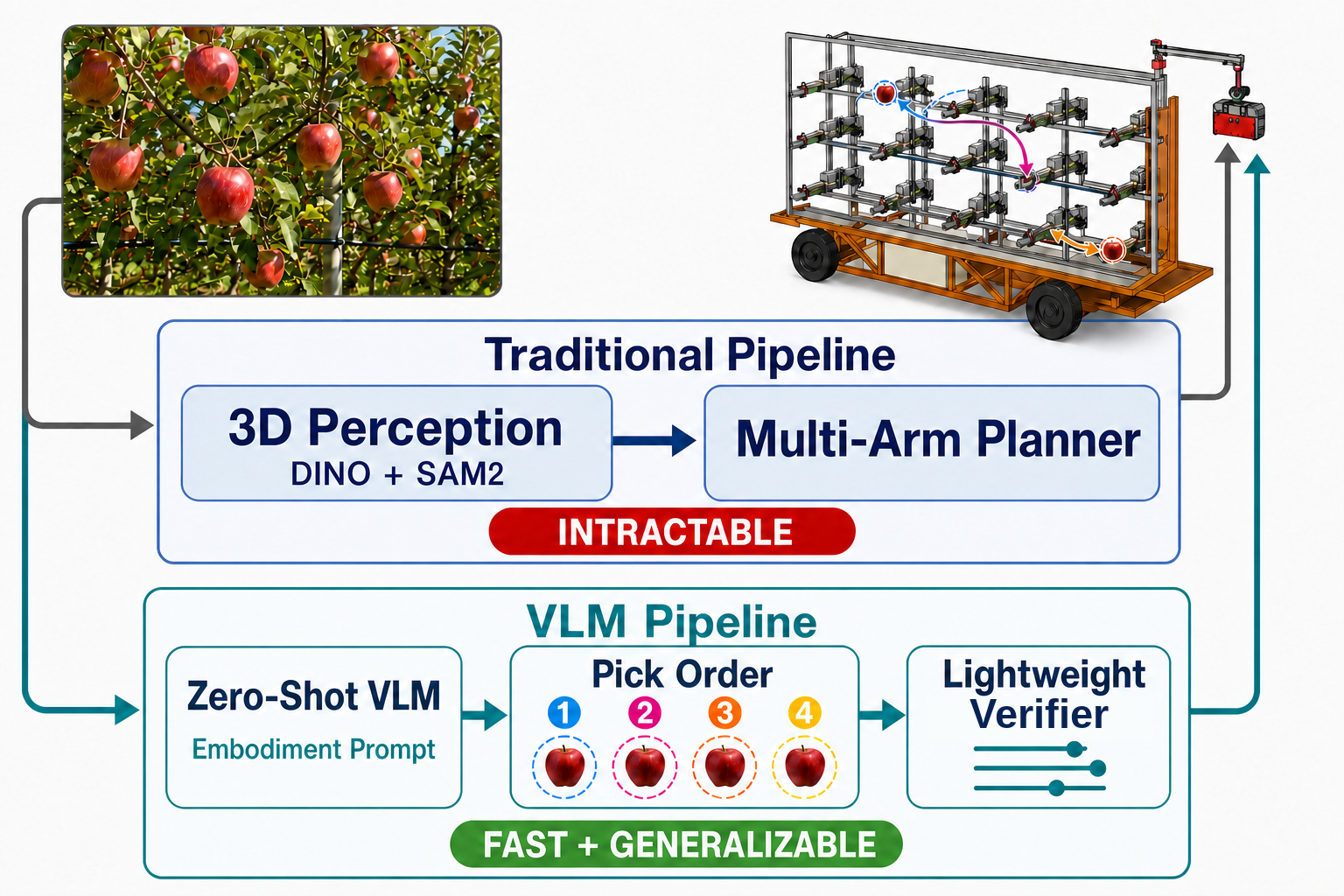}
\vspace{-6mm}
\caption{Overview of our benchmark for evaluating zero-shot VLMs in multi-arm robotic fruit harvesting. An oracle planner provides an upper-bound reference for assessing VLMs as a generalizable and efficient alternative to intractable multi-arm planning.}
\label{fig:intro}
\vspace{-6mm}
\end{figure}

%
Agricultural automation has become increasingly important as growers face persistent labor shortages and rising labor costs, particularly in labor-intensive specialty crop production. In the United States, labor accounts for approximately 40\% of production expenses for fruit and tree-nut farms~\cite{USDAFarmLabor}. Fruit harvesting is especially labor-intensive and time-sensitive, as delays can lead to unharvested fruit, reduced quality, and direct economic losses. Robotic harvesting therefore offers a promising solution to reduce reliance on seasonal manual labor and improve harvesting efficiency. In particular, compared with single-arm systems, \emph{multi-arm robotic harvesting systems} can increase throughput by picking fruits concurrently, providing a scalable path toward efficient and autonomous orchard production.

However, building a practical multi-arm robotic harvester that picks reliably, safely, and efficiently remains challenging. First, the system must generalize across crops and environments without extensive data collection and retraining. Traditional approaches rely on vision detectors such as YOLO~\cite{redmon2016yolo}, whose performance can degrade under changes in lighting, occlusion, and environmental conditions~\cite{arikapudi2016estimation,villacres2023apple}. Second, coordinating multiple arms in a shared workspace is computationally hard: each additional arm expands the trajectory-planning space, and the underlying harvest scheduling problem is NP-hard~\cite{garrett2021integrated}. Conventional optimization and heuristic methods manage this through simplifying assumptions, such as partitioning the workspace among arms~\cite{barnett2020work,zhu2025optimal}, which can reduce planning quality and harvesting throughput. Together, these challenges call for \emph{a more generalizable and efficient approach to multi-arm robotic harvesting.}

\textbf{Motivation.} Rather than solving the NP-hard harvesting problem optimally, human workers make effective decisions through visual understanding and reasoning under limited time and resources. Given an orchard scene such as the top-left of Fig.~\ref{fig:intro}, a worker can quickly infer the spatial relationships among fruits (e.g., which are in front or behind) and determine a practical harvesting sequence. This motivates us to explore \textbf{Vision-Language Models (VLMs)}, which exhibit aspects of the same human-like understanding and reasoning that can support complex planning tasks. Recent findings further suggest that VLMs are more effective at high-level task planning than precise low-level trajectory generation~\cite{zhao2025manipbench}, making them particularly well suited for generating harvesting sequences. Their prior knowledge and semantic reasoning may therefore enable a generalizable and efficient approach to multi-arm robotic harvesting, generating harvesting sequences for multiple arms within manageable computational resources. Despite this potential, to the best of our knowledge, no prior work has systematically evaluated the feasibility and limitations of pretrained VLMs for zero-shot multi-arm harvesting planning.

In this paper, we introduce the first comprehensive benchmark of VLMs for multi-arm robotic fruit harvesting planning. As shown in Fig.~\ref{fig:intro}, our benchmark takes real-world orchard images as input and evaluates planning outputs on a representative multi-arm Cartesian harvester, whose simple kinematics and parallel operation enable efficient picking in structured orchards~\cite{zhu2025fast}. As an oracle reference, we construct a traditional pipeline with separate 3D perception and planning stages, following state-of-the-art approaches~\cite{zhu2025foundation,zhu2025fast}. We then evaluate a zero-shot VLM pipeline that directly generates harvesting sequences and waypoints. We further introduce a lightweight trajectory verifier to measure collisions in VLM-generated plans, a critical safety requirement. Extending the benchmark to vision-language-action models and other embodiments is left for future work.

In summary, our contributions are as follows:
\begin{itemize}
\item We present the first comprehensive benchmark for evaluating pretrained VLMs on zero-shot multi-arm fruit harvesting planning. The benchmark covers two real-world crops, apple and citrus, and evaluates a broad set of recent closed-source and open-source VLMs.
\item We develop a VLM-based planning pipeline that generates multi-arm harvesting waypoints directly from raw orchard RGB-D images, together with a lightweight trajectory verifier for collision evaluation.
\item Our results show that frontier VLMs can generate effective zero-shot multi-arm harvesting plans when provided with physical-scale and workspace priors, although performance varies substantially across models, datasets, and scene complexity. Accurate 3D waypoint generation and collision-aware coordination remain key challenges.
\item We will open-source the benchmark upon the publication of this paper to facilitate future research in this direction.
\end{itemize}

\section{Related Work}
\label{sec:related_work}

\noindent
\textbf{Multi-Arm Robotic Fruit Harvesting.}
Existing approaches to multi-arm robotic fruit harvesting primarily rely on combinatorial optimization and heuristic algorithms~\cite{mann2016combinatorial,svoboda2024improving,zhu2025fast,zhu2025optimal}, or learning-based methods such as reinforcement learning~\cite{li2023multi,li2024intermittent}. Many existing systems partition the canopy into separate regions for individual arms, simplifying coordination but limiting harvesting throughput~\cite{barnett2020work,zhu2025optimal}. Recent heuristic schedulers improve efficiency for Cartesian multi-arm systems~\cite{zhu2025fast}, but rely on hand-crafted geometric assumptions that may not generalize to more flexible robotic embodiments.

These approaches also depend heavily on a separate perception pipeline. Their performance can degrade when fruit detection is affected by occlusion, poor illumination, or other environmental variations, directly reducing harvesting efficiency. Moreover, training robust fruit detectors typically requires substantial data collected from the target environment~\cite{arikapudi2016estimation,villacres2023apple}. Together, these limitations make existing multi-arm harvesting systems difficult to generalize and inefficient to deploy at scale.

\begin{figure*}[t]
    \centering
    \includegraphics[width=1\linewidth]{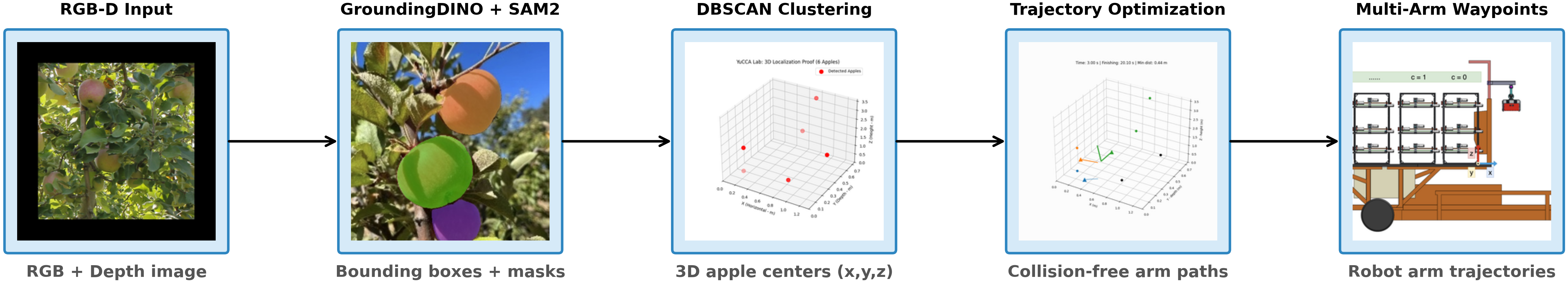}
    \vspace{-5mm}
    \caption{The three-stage oracle pipeline, consisting of open-vocabulary detection and segmentation, 3D target localization and task allocation, and multi-arm trajectory optimization. Each stage is implemented using state-of-the-art works for the corresponding task.}
    \label{fig:method}
    \vspace{-4mm}
\end{figure*}

\noindent
\textbf{VLM and VLA Benchmarks for Agriculture Robotics Tasks.}
Benchmarks are a key first step in foundation model development, providing a standardized and fair evaluation platform. Recent work has introduced VLM benchmarks for agricultural tasks such as crop and disease recognition~\cite{yang2025agrigpt,shinoda2025agrobench,li2025can,sapkota2025multi,boudiaf2026agrichat,gauba2025agmmu,arshad2025leveraging}. However, these efforts focus on perception and reasoning, with limited connection to downstream robotic actions.
In parallel, large-scale benchmarks for vision-language-action models~\cite{wang2025roboeval,zhang2025vlabench,liu2025eva,guruprasad2024benchmarking} and world action models~\cite{li2026causal,ye2026world} have begun to incorporate action. Yet, these methods are typically developed in controlled indoor settings using single-arm or bimanual platforms. In contrast, we focus on real-world fruit harvesting in unstructured orchard environments, where more than two arms are often needed to achieve high throughput.

\section{Problem Definition}
We consider multi-arm robotic fruit harvesting with a stationary harvesting vehicle equipped with $M$ Cartesian arms, following the representative system in~\cite{zhu2025fast}. Arms are mounted on a shared frame with uniform spacing $d$ and operate within an overlapping workspace in front of the canopy. A vehicle-mounted RGB-D camera captures a single-shot observation of the canopy and fruits within the reachable workspace before harvest begins. The planner, either traditional or VLM-based, generates a harvesting sequence and trajectory for each arm based on the observation and robot configuration. Each arm follows the Cartesian harvesting mechanism in~\cite{zhu2025fast}: it first moves laterally in the $x$-$z$ plane with the $y$-axis retracted to avoid interference with the canopy, then extends along the $y$-axis to approach and detach the target fruit after reaching its $x$-$z$ position.

Our benchmark evaluates the planner's ability to generate effective and collision-free multi-arm harvesting plans from this initial canopy observation.
In practice, additional cameras or sensors mounted on individual end effectors can provide closed-loop feedback and refine trajectories as the arms approach the fruits. We leave such online perception and replanning to future work. While our benchmark uses Cartesian arms as a representative embodiment, the formulation can be extended to other multi-arm robotic platforms.

\section{Benchmark Setup}

In this section, we describe the setup of our VLM benchmark for multi-arm robotic fruit harvesting. As shown in Fig.~\ref{fig:intro}, the benchmark consists of three key components: an oracle perception and planning pipeline (Sec.~\ref{sec:oracle}), a zero-shot VLM planning pipeline (Sec.~\ref{sec:vlm}), and a trajectory safety verifier (Sec.~\ref{sec:safety}).

\subsection{Oracle Perception and Planning Pipeline}
\label{sec:oracle}

To establish a strong and reproducible benchmark reference, we design the oracle pipeline following state-of-the-art perception and planning methods used in robotic fruit harvesting~\cite{zhu2025foundation,zhu2025fast}. As shown in Fig.~\ref{fig:method}, the pipeline converts a single RGB-D orchard image into collision-free multi-arm harvesting trajectories in three stages.

\noindent \textbf{Stage 1: Open-Vocabulary Detection and Segmentation.}
We adopt the perception pipeline from Zhu et al.~\cite{zhu2025foundation}. Given an RGB image, GroundingDINO~\cite{liu2023grounding} detects fruits from natural-language queries (e.g., ``ripe apples'') and returns their bounding boxes. We apply non-maximum suppression (NMS) with an IoU threshold $\tau_{IoU}$ and remove detections below a confidence threshold $\tau_{conf}$. For each remaining bounding box, SAM2~\cite{ravi2024sam2} generates a pixel-level fruit mask, separating the target fruit from the surrounding foliage and background. The corresponding depth pixels are then extracted for 3D localization.

\noindent \textbf{Stage 2: 3D Target Localization.}
We project the segmented depth pixels into 3D camera coordinates using the pinhole camera model:
\begin{equation}
X = \frac{(u-c_x)}{f}Z, \quad
Y = \frac{(v-c_y)}{f}Z, \quad
Z = d(u,v),
\end{equation}
where $(u,v)$ denotes the pixel coordinate, $f$ is the focal length, $(c_x,c_y)$ is the principal point, and $d(u,v)$ is the measured depth. Depth values outside the valid range $[d_{\min},d_{\max}]$ are discarded to reduce measurement noise.

We then apply DBSCAN~\cite{ester1996density} to group the projected points into individual fruit instances. With neighborhood radius $\epsilon$ and minimum density threshold $n_{\min}$, it suppresses sparse depth noise and recovers compact clusters, whose centers give the estimated 3D fruit positions $(x,y,z)$. Each fruit is also assigned a \textit{detach time}, the time an arm needs to detach it from the branch. Finally, we define the robotic workspace by padding the spatial extent of the detected fruits by a fixed $\gamma$.

\noindent \textbf{Stage 3: Task Allocation and Multi-Arm Trajectory Optimization.}
The estimated 3D fruit positions are passed to the state-of-the-art bi-level mixed-integer programming (MIP) framework in~\cite{zhu2025fast}, which jointly addresses task allocation and trajectory planning. At a high level, the framework first assigns fruits to individual arms and then generates synchronized, collision-free trajectories for all arms. 


\begin{figure*}[t]
\centering
\begin{tcolorbox}[
  colback=gray!10,
  colframe=gray!50,
  title=VLM Prompt
]
\begin{verbatim}
ROLE: You are an expert Agricultural Robotics Vision System specialized in 3D
fruit localization and path planning. Your goal is to identify ripe fruit and
generate collision-free robotic picking waypoints with millimeter-level 
precision.
### PHYSICAL SCALE REFERENCE:
- A ripe {FRUIT} in this image is 6-10 cm (0.06-0.10 m) wide.
- Use this to calibrate the workspace (likely 1.0m to 5.0m).
### COORDINATE ALIGNMENT:
- CAMERA CENTER is (0.0, 0.0, 0.0).
- Coordinates X (Horizontal) and Z (Vertical) are relative to center 0.0.
- Depth (Y) is absolute meters from camera (approx 2.5m-3.5m).
{DEPTH}
### ROBOT: {N} Cartesian arms on a shared rail, ordered left to right along X.
Arms cannot pass each other: every fruit on Arm_k must have a smaller X than
every fruit on Arm_k+1, or they collide.
### TASK:
1. Define "Harvesting Zone" boundaries based on visible {FRUIT}.
2. {ZONE}
3. Generate picking waypoints [x, y, z] in METERS for each assigned volume.
Output ONLY a JSON object:
{
  "workspace_metadata": { "detected_width_m": float, "num_arms": {N} },
  "picking_order": { "Arm_1": [], ..., "Arm_{N}": [] },
  "trajectory_steps_meters": { "Arm_1": [[x,y,z]], ..., "Arm_{N}": [] },
  "reasoning": "Explain the scaling and partitioning logic."
}
\end{verbatim}
\end{tcolorbox}
\vspace{-5mm}
\caption{Prompt template used for zero-shot VLM-based multi-arm harvesting planning. 
The placeholders \texttt{\{FRUIT\}}, \texttt{\{N\}}, \texttt{\{DEPTH\}}, and 
\texttt{\{ZONE\}} are instantiated according to the crop, robot configuration, 
and experimental setting.}
\label{fig:vlm_prompt}
\end{figure*}

\subsection{Zero-Shot VLM Planning Pipeline}
\label{sec:vlm}

Our goal is to evaluate whether pretrained VLMs can generalize multi-arm harvesting planning across crops and orchard environments without retraining.
Therefore, the key design choice in this pipeline is the prompt: it must provide sufficient physical and robotic context for planning while relying on the VLM's pretrained visual understanding and reasoning capabilities. We use the same prompt structure across all VLMs to ensure a fair comparison.

\noindent \textbf{Prompt Design.}
Fig.~\ref{fig:vlm_prompt} shows the prompt template used across all VLMs.
The VLM receives the raw orchard image together with basic information about the robotic workspace and physical constraints. Rather than providing pre-detected fruit coordinates, we ask the VLM to visually identify the fruits, reason about their spatial relationships, and directly generate a harvesting sequence and waypoints for each arm. The prompt also specifies the reachable workspace of each arm to guide task allocation and reduce infeasible assignments. Finally, we require a structured JSON output that can be automatically parsed and evaluated by our benchmark.

For each run, four prompt placeholders are instantiated to specify the crop, robot configuration, and information provided to the model, while keeping the overall prompt structure fixed, as shown in Fig.~\ref{fig:vlm_prompt}:

\begin{itemize}
\renewcommand{\labelitemi}{--}
\item \texttt{{FRUIT}} specifies the target crop, either apple or citrus. The same crop definition is used in the oracle pipeline, ensuring a consistent target across both pipelines.

\item \texttt{{N}} specifies the number of Cartesian arms and is varied to evaluate scalability as multi-arm coordination becomes more challenging.

\item \texttt{{DEPTH}} specifies whether depth information is provided alongside the RGB image, allowing us to evaluate the contribution of explicit depth cues.

\item \texttt{{ZONE}} specifies how the workspace is partitioned among the arms. We consider four strategies: (i) no partitioning, (ii) equal segments along the $x$-axis, (iii) equal bands along the $z$-axis, and (iv) a grid over both axes. These settings allow us to evaluate how workspace partitioning affects fruit allocation and planning quality. By default, we use no partitioning to give the planner maximum flexibility in coordinating the arms.

\end{itemize}

\noindent Our prompt further incorporates three key design elements:

\begin{itemize}
\renewcommand{\labelitemi}{--}
\item \textbf{Physical Scale:}
We provide a physical reference to help the VLM map image observations to real-world coordinates. The prompt specifies a fruit diameter range of $6$--$10$ cm rather than a single value, covering a broader range of crops without implying unrealistic precision from a single view. This allows the model to estimate scale and generate waypoints in metric units.

\item \textbf{Workspace Constraints:} We provide the dimensions of the robotic workspace and the reachable region of each arm, together with the ordering constraint imposed by the shared frame. This information guides the VLM in assigning fruits to feasible arms and generating trajectories within the physical workspace.

\item \textbf{Structured Output:}
We require all VLMs to follow a fixed JSON schema containing the harvesting order and waypoints for each arm. We also request a brief explanation of the model's spatial and scaling reasoning for failure analysis. This format enables automatic parsing and consistent evaluation across models.

\end{itemize}

\subsection{Trajectory Verification and Safety Evaluation}
\label{sec:safety}

Collisions are a critical safety concern in multi-arm harvesting, where several arms operate simultaneously within a shared workspace. A harvesting plan is therefore only useful if the arms can execute it without collision. Verifying this is challenging because the VLM outputs waypoints but no timing information: we know the order in which each arm visits its assigned fruits, but not when it arrives. Rather than searching over all possible schedules, we exploit the kinematic structure of the Cartesian harvester to derive a timing-independent feasibility check. The $M$ arms are mounted on a shared frame and remain ordered along the $x$-axis, so arm $i$ cannot pass arm $i{+}1$. Let $\mathcal{A}_i$ denote the fruits assigned to arm $i$. If there exist $p \in \mathcal{A}i$ and $q \in \mathcal{A}{i+1}$ such that $x_p > x_q$, the two arms would need to exchange their relative order during execution, which is mechanically infeasible. We refer to this condition as an \emph{order violation}. The verifier only checks the generated plan and does not modify or repair it. In our benchmark, plans containing an order violation are recorded as collisions, and we report the collision ratio as the percentage of evaluated scenes containing at least one such violation.
\section{Results}
\label{sec:results}

\subsection{Experimental Setup}

\noindent\textbf{Datasets.} We evaluate on two real-world orchard datasets. For
apples we use a curated version of the Fuji RGB-D dataset~\cite{zhukeyi2026fuji}, which
provides registered RGB and depth images with manually annotated fruit
instances. For citrus we use~\cite{hou2022citrus}, comprising 432 RGB images with depth
maps and fruit annotations. From each dataset we hold out a benchmark set of
scenes spanning a range of fruit counts, lighting conditions, and occlusion
levels. All VLMs are evaluated on the identical scene set per crop.

\noindent\textbf{Models.} We evaluate five closed-source and two open-source
VLMs, listed in Table~\ref{tab:vlm_models}.
\begin{table}[t]
    \centering
    \small
    \caption{VLMs evaluated in our benchmark.}
    \vspace{-3mm}
    \label{tab:vlm_models}
    \begin{tabularx}{\linewidth}{@{}lX@{}}
        \toprule
        \textbf{Category} & \textbf{Models} \\
        \midrule
        Closed-source
        & GPT-4o~\cite{openai2024gpt4o}, GPT-5.5-Pro~\cite{openai2025gpt5},
        \mbox{Gemini 3 Flash}~\cite{google2025gemini3}, Claude Sonnet 3.5~\cite{anthropic2024claude},
        \mbox{Claude Opus 4.7}~\cite{anthropic2024claude} \\
        \hline 
        Open-source
        & Qwen2-VL-7B~\cite{qwen25vl2025}, GLM 4.6~\cite{glm45v2025} \\
        \bottomrule
    \end{tabularx}
\vspace{-5mm}
\end{table}
All models receive the same instantiated
prompt for a given condition and are queried zero-shot, without fine-tuning.

\noindent\textbf{Metrics.}
We evaluate each VLM using five metrics spanning harvesting performance, safety, and efficiency. The \textit{detection ratio} measures the percentage of ground-truth fruits identified by the model within a loose spatial threshold, indicating whether the model successfully recognizes and locates each fruit. The \textit{harvested ratio} measures the percentage of ground-truth fruits for which the generated waypoint lies within the selected distance threshold, indicating whether the arm is guided sufficiently close to the fruit for harvesting. The \textit{collision ratio} measures the percentage of scenes containing an infeasible arm assignment, as defined in Sec.~\ref{sec:safety}. For efficiency, we report the total \textit{trajectory distance} traveled by all arms and the average number of \textit{tokens} used per scene.
\noindent We experiment on a workstation equipped with three NVIDIA RTX PRO 6000 GPUs~\cite{nvidia2022rtx6000ada}. 

\noindent\textbf{Oracle pipeline Parameters.} We follow the perception protocol of Zhu et
al.~\cite{zhu2025foundation}, with NMS IoU threshold $\tau_{IoU}=0.789$ and
detection confidence $\tau_{conf}=0.3$. DBSCAN uses a neighborhood radius
$\epsilon=0.04$\,m, set to half a standard fruit diameter so that adjacent
fruits are not merged into a single instance, with minimum density
$n_{\min}=10$.

\begin{figure*}[h!]
    \centering
    \includegraphics[width=0.99\textwidth]{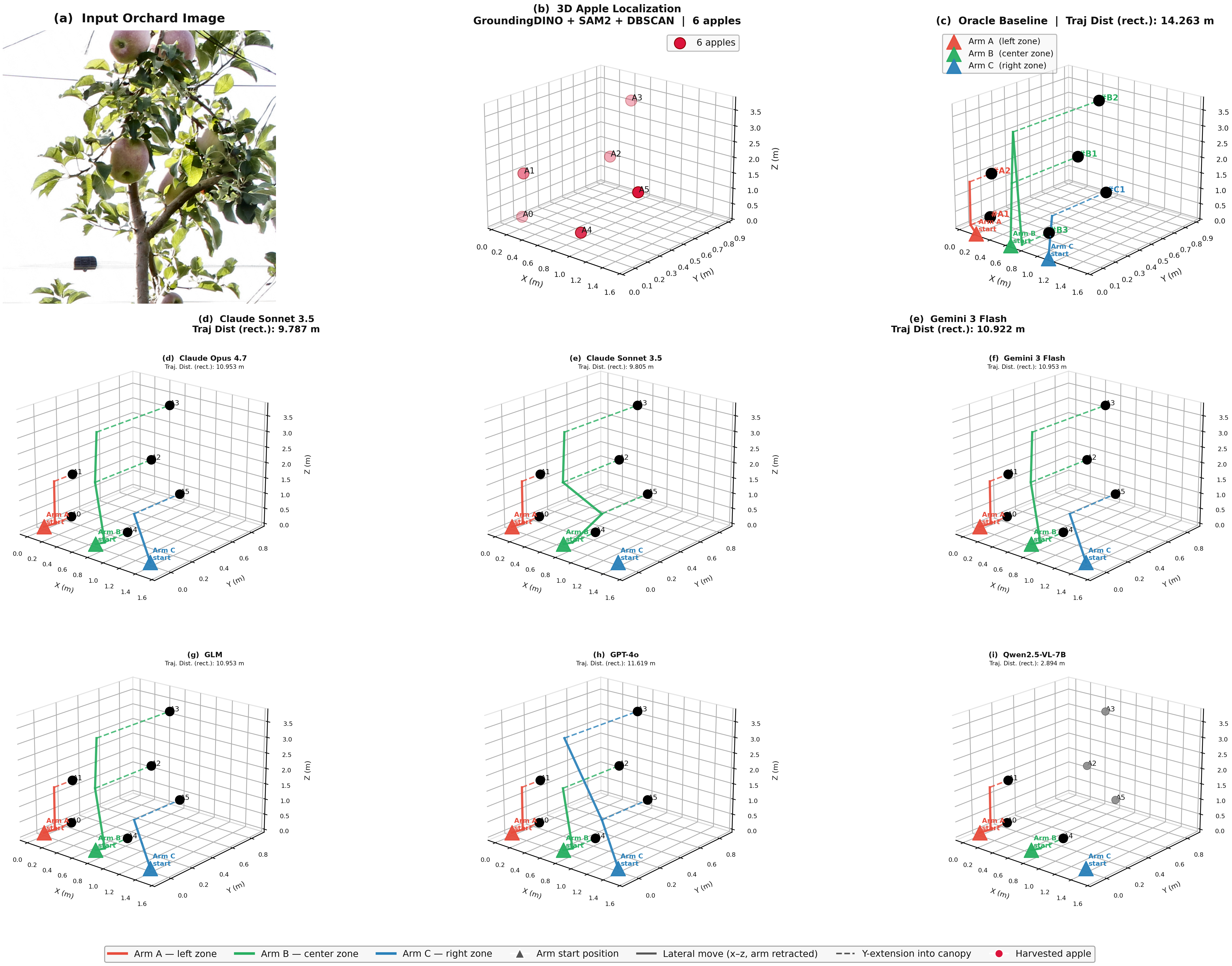}
    \vspace{-3mm}
    \caption{Qualitative comparison of VLM-generated trajectories against the oracle
  pipeline on a representative 6-fruit orchard scene using a 3-arm robot.
  \textbf{(a)} Input RGB orchard image.
  \textbf{(b)} 3D fruit localization from the oracle pipeline.
  \textbf{(c)} Oracle trajectory, reported with rectilinear trajectory distance.
  \textbf{(d--i)} VLM-generated trajectories for Claude Opus~4.7, Claude
  Sonnet~3.5, Gemini~3 Flash, GLM, GPT-4o, and Qwen2.5-VL-7B, respectively.
  Each panel reports its rectilinear trajectory distance. Red and grey fruit
  markers denote fruit included and not included, respectively, in the model's
  harvesting plan.}
\label{fig:qualitative}
\vspace{-4mm}
\end{figure*}

\subsection{Qualitative Results}
Figure~\ref{fig:qualitative} compares VLM-generated harvesting trajectories with the oracle baseline on a representative 6-apple orchard scene using a 3-arm robot.

\noindent\textbf{Most frontier VLMs generate complete harvesting plans.}
The oracle baseline includes all six fruits in its plan. Claude Opus 4.7, Claude Sonnet 3.5, Gemini 3 Flash, GLM, and GPT-4o similarly achieve full target coverage. In contrast, Qwen2.5-VL-7B omits several fruits, shown as gray markers, resulting in an incomplete plan. This highlights a clear gap in plan completeness between stronger frontier models and the smaller open-source model.

\noindent\textbf{Trajectory efficiency varies substantially across models.}
Despite receiving the same scene and robot configuration, the models produce markedly different path lengths. Claude Sonnet 3.5 (9.7 m), Gemini 3 Flash (10.0 m), and GPT-4o (10.4 m) generate shorter trajectories than the oracle baseline (14.2 m), while GLM produces a considerably longer path of 20.0 m. These results show that successful fruit coverage does not necessarily imply efficient motion planning.

\noindent\textbf{Spatial allocation remains an important source of variation.}
The models must distribute fruits across the three arm workspaces while maintaining feasible trajectories. Stronger models generally produce compact and well-structured routes, whereas less effective models incur unnecessary travel when assigning fruits near workspace boundaries. Overall, the qualitative results suggest that frontier VLMs can generate complete and efficient multi-arm plans, but their performance remains sensitive to fruit allocation and waypoint selection.

\begin{table*}[t]
\small
\centering
\caption{ Comprehensive performance comparison of VLM backends for multi-arm fruit-harvesting planning on the Apple and Citrus datasets. Metrics include Harvested Ratio across three spatial thresholds ($\delta$), collision ratio, trajectory distance, and average token usage. Results are benchmarked against the optimal planner. Best reported VLM result in each column is shown in \textcolor{red}{Red text} for apples and \textcolor{orange}{Orange text} for citrus.}
\label{tab:vlm_master_results}
\vspace{-3mm}
\begin{tabular}{llccccccc}
\toprule
\textbf{Model} & \textbf{Dataset}
& \multicolumn{3}{c}{\textbf{Harvested Ratio (\%)$\uparrow$}}
& \textbf{Detected}
& \textbf{Collision}
& \textbf{Traj.}
& \textbf{Avg.} \\
\cmidrule(lr){3-5}
&
& \textbf{$\delta=1.0$m}
& \textbf{$\delta=0.5$m}
& \textbf{$\delta=0.25$m}
& \textbf{Ratio (\%)$\uparrow$}
& \textbf{Ratio (\%)$\downarrow$}
& \textbf{Dist. (m)$\downarrow$}
& \textbf{Tokens$\downarrow$} \\
\midrule

\multirow{2}{*}{Optimal Planner$^{*}$}
& Apple
& 100.0 & 100.0 & 100.0 & 100.0 & 100.0 & 0 & 12.4  \\
& Citrus
& 100.0 & 100.0 & 100.0 & 100.0 & 100.0 & 0 & 5.5 \\

\midrule
\multicolumn{9}{l}{\textit{Closed-source VLMs}} \\

\multirow{2}{*}{GPT-5.5-Pro \cite{openai2025gpt5}}
& Apple
& 93.8 & \textcolor{red}{85.7} & 42.4 & 93.8 & 42.2 & 12.8 & 6364 \\
& Citrus
& 99.3 & 60.7 & 6.7 & 89.8 & 37.4 & 13.7 & 8175 \\

\multirow{2}{*}{Gemini 3 Flash \cite{google2025gemini3}}
& Apple & 79.2 & 41.2 & 89.8 & \textcolor{red}{98.2} & 48.6 & 14.3 & \textcolor{red}{1150} \\
& Citrus
& 65.9 & 31.7
& 97.6 & 97.6 & 35.5 & 7.9 & 7104 \\

\multirow{2}{*}{Claude Sonnet 3.5 \cite{anthropic2024claude}}
& Apple
& \textcolor{red}{98.2} & 65.3 & 21.8 & 81.0 & 86.4 & 20.5 & 1200 \\
& Citrus
& 90.3 & 67.7 & 41.9 & 90.3 & 82.8 & 11.8 & 3088 \\

\multirow{2}{*}{Claude Opus 4.7 \cite{anthropic2025opus}}
& Apple
& 81.0 & 69.5 & 27.4 & 81.0 & 38.1 & 12.7 & 1527 \\
& Citrus
& \textcolor{orange}{98.8} & 53.7 & 8.5 & \textcolor{orange}{98.8} & 31.3 & 8.4 & 2113 \\

\multirow{2}{*}{GPT-4o \cite{openai2024gpt4o}}
& Apple
& 41.9 & 20.4 & 2.1 & 41.9 & 12.1 & 11.1 & 1314 \\
& Citrus
& 60.5 & 20.9 & 9.3 & 60.5 & \textcolor{orange}{4.3} & 6.2 & \textcolor{orange}{2103} \\

\midrule
\multicolumn{9}{l}{\textit{Open-source VLMs}} \\

\multirow{2}{*}{Qwen2.5-VL-7B \cite{qwen25vl2025}}
& Apple
& 60.5 & 60.5 & 37.7 & 60.5 & \textcolor{red}{11.9} & 10.9 & 9553 \\
& Citrus
& 98.7 & 87.7 & 36.1 & 98.7 & 31.8 & \textcolor{red}{2.3} & 8226 \\

\multirow{2}{*}{GLM \cite{glm45v2025}}
& Apple
& 86.9 & 85.0 & \textcolor{red}{64.5} & 86.9 & 17.9 & 11.8 & 2454 \\
& Citrus
& 90.3 & \textcolor{orange}{89.6} & \textcolor{orange}{88.9} & 82.9 & 25.0 & \textcolor{orange}{2.8} & 2564 \\

\bottomrule
\end{tabular}

\vspace{2pt}
\begin{minipage}{0.98\textwidth}
\footnotesize
$^{*}$The optimal planner assumes oracle knowledge of the exact 3D positions of all fruits. It therefore serves as an upper-bound reference and achieves a 100\% harvested ratio by considering all reachable fruits for harvesting. Collision ratio was not recorded for the current VLM evaluations; `--` denotes an unmeasured value. The Apple Qwen row uses 48 common test scenes and the Citrus Qwen row uses 44 scenes; both are direct RGB-only waypoint predictions. Citrus backend rows use the available static-planner evaluations and therefore do not all have identical scene counts.
\end{minipage}
\vspace{-4mm}
\end{table*}

\subsection{Quantitative Results}
\label{sec}

Table~\ref{tab:vlm_master_results} summarizes the quantitative performance of all evaluated VLMs. Trajectory accuracy is evaluated at three spatial thresholds, $\delta \in \{1.0\text{m}, 0.5\text{m}, 0.25\text{m}\}$, where $\delta$ denotes the maximum Euclidean distance between a VLM-generated waypoint and the corresponding ground-truth fruit location.

\paragraph{Performance Analysis.}
Performance varies substantially across models, datasets, and spatial thresholds. On Apple, GPT-5.5-Pro achieves the highest closed-source accuracy at $\delta=0.5$ m (85.7\%), while GLM performs best at the strictest threshold of $\delta=0.25$ m (64.5\%). On Citrus, GLM achieves the strongest performance at both $\delta=0.5$ m and $\delta=0.25$ m, reaching 89.6\% and 88.9\%, respectively.

High accuracy at a loose threshold, however, does not necessarily translate to precise and safe harvesting. On Citrus, GPT-5.5-Pro achieves 99.3\% harvested ratio at $\delta=1.0$ m but only 6.7\% at $\delta=0.25$ m. Similarly, Claude Sonnet 3.5 achieves the highest Apple harvested ratio at $\delta=1.0$m (98.2\%) but also exhibits the highest collision ratios on Apple (86.4\%) and Citrus (82.8\%). These results show that coarse fruit localization alone is insufficient for reliable multi-arm harvesting.

\noindent\textbf{Sensitivity to Spatial Granularity.}
Performance generally decreases as the spatial threshold becomes stricter, although the degree of degradation varies considerably across models and datasets. GPT-4o degrades sharply at tighter thresholds, while GPT-5.5-Pro exhibits a particularly large gap between coarse and precise localization on Citrus. In contrast, GLM remains comparatively robust at stricter thresholds, especially on Citrus. Overall, the results highlight the importance of precise metric localization beyond simply identifying the approximate location of each fruit.

\noindent\textbf{Inference Efficiency.}
Strong planning performance can also come at substantial computational cost. GPT-5.5-Pro averages 6,364 tokens per Apple scene and 8,175 tokens per Citrus scene, considerably more than several alternatives. This overhead may limit its practicality for large-scale or real-time deployment and highlights the importance of balancing planning quality with inference efficiency.

\subsection{Failure Case Analysis}

We further analyze VLM outputs against the oracle fruit locations and multi-arm plans. The dominant failure modes arise from incomplete fruit detection, inaccurate metric localization, and infeasible multi-arm coordination.

\noindent\textbf{Perception Failures.}
Dense foliage, occlusion, and overlapping fruits can cause VLMs to miss instances, resulting in incomplete harvesting plans. Such failures occur more frequently in crowded scenes and near image boundaries.

\noindent\textbf{Spatial Localization Failures.}
VLMs can often estimate approximate fruit locations, but precise metric localization directly from RGB images remains challenging. Errors are particularly pronounced along the depth ($Y$) direction, where metric distance is difficult to infer from monocular visual cues. Consequently, some models perform well at $\delta=1.0$ m but degrade substantially at $\delta=0.5$ m and $\delta=0.25$ m.

\noindent\textbf{Multi-Arm Planning Failures.}
Even when fruits are correctly detected, a VLM may assign them to unsuitable arms or generate trajectories that violate safety constraints. Such coordination failures become more frequent as the number of arms increases, highlighting the additional difficulty of scaling from perception to multi-arm planning.

\subsection{Prompt Choice Analysis}

To analyze prompt component sensitivity, we evaluate six ablation variants across all models by systematically removing elements from the full prompt (V1):

\begin{itemize}
    \item \textbf{V1 Full}: Role framing, physical scale reference, coordinate frame, and structured JSON output.
    \item \textbf{V2 w/o Scale}: V1 without the 8\,cm apple size prior.
    \item \textbf{V3 w/o Role}: V1 without expert agricultural robotics role assignment.
    \item \textbf{V4 Minimal}: Task description and JSON schema only.
    \item \textbf{V5 w/o Depth}: V1 without the expert agricultural robotics role assignment. 
    \item \textbf{V6 Zone Splitting}: V1 without the expert agricultural robotics role assignment.
\end{itemize}

Table~\ref{tab:prompt_ablation_summary} reports mean and standard deviation across models to isolate prompt component effects.
\paragraph{Physical scale is critical.}
Removing the 8\,cm size prior (\textbf{V2}) causes the largest drop in localization accuracy, reducing mean harvested ratio at $\delta=0.5$\,m from $73.5\%$ to $47.9\%$ and at $\delta=0.25$\,m from $36.9\%$ to $22.3\%$. Target detection remains high ($80.8\%$), showing that scale is necessary for precise 3D waypoint placement rather than fruit recognition.
\paragraph{Scale and role priors buy precision, not recall.}
Removing either the fruit-size prior (\textbf{V2}) or the expert role
(\textbf{V3}) leaves detection close to the full prompt while harvested ratio
falls by roughly a third at every threshold. Both help a model place fruit
accurately rather than find it, and their wide variance shows some models carry
these priors internally while others depend on the prompt.

\paragraph{Structured grounding is the one component nothing survives without.}
The minimal prompt (\textbf{V4}) harvests nothing at any threshold while still
detecting nearly half the fruit. The models see the orchard but cannot express
it as something a planner can execute. No other ablation fails categorically
like this.

\paragraph{Depth anchoring splits the models; zone splitting is redundant.}
Removing the depth statement (\textbf{V5}) produces the widest variance in the
study, so its mean hides two populations: models that recover metric depth
unaided and models that fail outright. Removing zone splitting (\textbf{V6})
barely registers, and its performance holds as the threshold tightens while
every other variant loses close to half. Models reconstruct the workspace
partition on their own.
\begin{table}[t]
\centering
\setlength{\tabcolsep}{3pt}
\renewcommand{\arraystretch}{1.1}
\caption{Prompt ablation summary, averaged across all five VLMs (mean $\pm$ std). Each variant removes one component from the full prompt (V1). Best results are shown in \textbf{bold}.}
\label{tab:prompt_ablation_summary}
\resizebox{\columnwidth}{!}{%
\begin{tabular}{l ccc c}
\toprule
\textbf{Prompt} 
& \multicolumn{3}{c}{\textbf{Harvested Ratio (\%)}}
& \textbf{Detected Ratio} \\
\cmidrule(lr){2-4}
& $\delta{=}1.0$m & $\delta{=}0.5$m & $\delta{=}0.25$m & \textbf{(\%)} \\
\midrule
V1 -- Full           & 93.3 $\pm$ 14.9 & 73.5 $\pm$ 22.4 & 36.9 $\pm$ 14.9 & 84.4 $\pm$ 8.4 \\
V2 -- w/o Scale      & 74.7 $\pm$ 35.4 & 47.9 $\pm$ 44.9 & 22.3 $\pm$ 24.8 & 80.8 $\pm$ 14.9 \\
V3 -- w/o Role       & 75.7 $\pm$ 33.6 & 43.2 $\pm$ 32.5 & 16.2 $\pm$ 16.4 & 75.9 $\pm$ 18.0 \\
V4 -- Minimal        & 0.0 $\pm$ 0.0   & 0.0 $\pm$ 0.0   & 0.0 $\pm$ 0.0   & 45.3 $\pm$ 33.3 \\
V5 -- w/o Depth      & 96.9 $\pm$ 5.4 & 86.5 $\pm$ 23.4 & 49.0 $\pm$ 44.2 & 96.9 $\pm$ 5.4 \\
V6 -- w/o Zone       & \textbf{98.7 $\pm$ 0.8} & \textbf{94.7 $\pm$ 1.1} & \textbf{89.3 $\pm$ 1.6} & \textbf{89.3 $\pm$ 1.6} \\
\bottomrule
\end{tabular}%
}
\end{table}
\FloatBarrier

\subsection{Scalability Analysis}

\noindent\textbf{Impact of the Number of Fruits.} 
Table~\ref{tab:scale_apples} shows the results of GPT-5.5-Pro under a fixed 2-arm configuration using scenes grouped by fruit density ($<5$, $5$--$10$, and $>10$ fruits per image). On Apple, harvesting accuracy drops significantly as fruit count increases, falling from $60.1\%$ to $42.8\%$. This confirms that denser, higher-complexity scenes negatively affect both spatial localization and target resolution. For Citrus, harvesting accuracy remains bounded below $62.7\%$ across all fruit density groups.

\begin{table}[t]
\centering
\small
\caption{Impact of fruit count on GPT-5.5-Pro performance with 2 arms fixed. $n$ denotes the number of evaluated scenes in each group. Apple performance decreases as scene complexity increases, whereas Citrus detection remains near-perfect and harvested ratio varies
  across count groups.}
\label{tab:scale_apples}
\resizebox{\linewidth}{!}{%
\begin{tabular}{llccc}
\toprule
\textbf{Dataset} & \textbf{Fruit Count} & \textbf{$n$ (scenes)} & \textbf{Detection Ratio (\%)} & \textbf{Harvested Ratio (\%)} \\
\midrule
\multirow{3}{*}{Apple}
& $< 5$ apples & 43 & \textbf{40.1} & \textbf{60.1} \\
& $5$--$10$ apples & 86 & 38.1 & 53.5 \\
& $> 10$ apples & 5 & 23.8 & 42.8 \\
\midrule
\multirow{3}{*}{Citrus}
& $< 5$ fruits & 46 & \textbf{100.0} & 47.0 \\
& $5$--$10$ fruits & 56 & 99.2 & \textbf{62.7} \\
& $> 10$ fruits & 2 & 100.0 & 52.2 \\
\bottomrule
\end{tabular}%
}
\end{table}

\noindent\textbf{Impact of the Number of Arms.} 
Table~\ref{tab:scale_arms} shows the result of GPT-5.5-Pro across 2-, 3-, and 4-arm configurations while maintaining a fixed scene complexity of $5$--$10$ fruits per image. As the arm count increases, overall harvesting success degrades from $52.1\%$ down to $23.8\%$ on Apple and from $44.3\%$ down to $23.6\%$ on Citrus, representing relative performance drops of $54\%$ and $47\%$, respectively. 

Scaling the number of arms expands the spatial coordination search space, making collision-free trajectory planning and dynamic zone partition significantly more challenging even when fruit count remains constant.
\begin{table}[t]
\centering
\small
\caption{Impact of arm count on GPT-5.5-Pro performance for scenes containing 5--10 apples. $n$ denotes the number of evaluated scenes per configuration. Both detection and harvested ratios decrease as the number of arms increases.}
\label{tab:scale_arms}
\resizebox{\linewidth}{!}{%
\begin{tabular}{llccc}
\toprule
\textbf{Dataset} & \textbf{Arm Count} & \textbf{$n$ (scenes)} & \textbf{Detection Ratio (\%)} & \textbf{Harvested Ratio (\%)} \\
\midrule
\multirow{3}{*}{Apple}
& 2 arms & 50 & 52.1 & 57.1 \\
& 3 arms & 50 & 34.2 & 55.7 \\
& 4 arms & 34 & 23.8 & 51.8 \\
\midrule
\multirow{3}{*}{Citrus}
& 2 arms & 45 & 44.3 & 53.8 \\
& 3 arms & 45 & 36.5 & 51.0 \\
& 4 arms & 45 & 23.6 & 48.6 \\
\bottomrule
\end{tabular}%
}
\end{table}
\section{Conclusion}
Multi-arm robotic fruit harvesting has the potential to improve harvesting efficiency, but practical deployment remains challenging due to the need for robust perception and coordinated planning in unstructured orchard environments. Existing methods often rely on task-specific perception pipelines or simplifying planning assumptions, limiting their generalizability. In this work, we introduced the first comprehensive benchmark for evaluating pretrained VLMs on zero-shot multi-arm fruit harvesting planning using real-world apple and citrus orchard images. Our results show that frontier VLMs can generate effective multi-arm harvesting plans without task-specific training, particularly when provided with physical-scale and workspace priors. However, accurate 3D waypoint generation and collision-aware coordination remain key challenges for practical deployment. We hope this benchmark provides a foundation for future research on generalizable and efficient VLM-based planning for multi-arm agricultural robotics.

\section*{Acknowledgements}
This research was supported by the Academic Senate Faculty Research Grant and the startup funds at University of California Merced. V. Inukollu and A. Zaman are supported by the Workforce Innovation Program (WIP) at the Center for Information Technology Research in the Interest of Society and the Banatao Institute (CITRIS).
{
    \small
    \bibliographystyle{ieeenat_fullname}
    \bibliography{main}
}

\end{document}